\def\year{2022}\relax
\documentclass[letterpaper]{article} 
\usepackage{aaai22}  
\usepackage{times}  
\usepackage{helvet}  
\usepackage{courier}  
\usepackage[hyphens]{url}  
\usepackage{graphicx} 
\usepackage{natbib}  
\usepackage{caption} 
\DeclareCaptionStyle{ruled}{labelfont=normalfont,labelsep=colon,strut=off} 
\usepackage{algorithm}
\usepackage{algorithmic}
\usepackage[utf8]{inputenc} 
\usepackage[T1]{fontenc}    
\usepackage{hyperref}       
\usepackage{url}            
\usepackage{booktabs}       
\usepackage{amsfonts}       
\usepackage{nicefrac}       
\usepackage{microtype}      
\usepackage{graphicx}
\usepackage{amsthm}
\usepackage{booktabs}
\usepackage{algorithm}
\usepackage{algorithmic}
\usepackage{amsfonts}
\usepackage{amsmath, bm}
\usepackage{multirow}
\usepackage{bbm}
\usepackage{color}
\usepackage{xcolor}
\usepackage{caption}
\usepackage{subcaption}
\usepackage{graphicx}
\usepackage[toc,page]{appendix}
\usepackage{hyperref}
\usepackage{array}
\usepackage{xhfill}
\usepackage{mathtools,tikz,caption}
\newcolumntype{L}{>{\centering\arraybackslash}m{3cm}}

\newtheorem{definition}{Definition}

\DeclareRobustCommand\sampleline[1]{%
  \tikz\draw[#1] (0,0) (0,\the\dimexpr\fontdimen22\textfont2\relax)
  -- (1.5em,\the\dimexpr\fontdimen22\textfont2\relax);%
}

\usepackage{newfloat}
\usepackage{listings}
\floatstyle{ruled}
\newfloat{listing}{tb}{lst}{}
\floatname{listing}{Listing}
\nocopyright

\title{Poisoning Attacks on Fair Machine Learning}

\author {
    Minh-Hao Van,\textsuperscript{\rm 1}
    Wei Du, \textsuperscript{\rm 1}
    Xintao Wu, \textsuperscript{\rm 1}
    Aidong Lu \textsuperscript{\rm 2}
}
\affiliations {
    \textsuperscript{\rm 1} University of Arkansas at Fayetteville\\
    \textsuperscript{\rm 2} University of North Carolina at Charlotte\\
    haovan@uark.edu, wd005@uark.edu, xintaowu@uark.edu, alu1@uncc.edu
}

\usepackage{bibentry}
\begin{document}
\maketitle
\begin{abstract}
Both fair machine learning and adversarial learning have been extensively studied.  However, attacking fair machine learning models has received less attention. In this paper, we present a framework that seeks to effectively generate poisoning samples to attack both model accuracy and algorithmic fairness. Our attacking framework can target fair machine learning models trained with a variety of group based fairness notions such as demographic parity and equalized odds.  We develop three online attacks, adversarial sampling , adversarial labeling,  and adversarial feature modification. All three attacks effectively and efficiently produce poisoning samples via sampling, labeling, or modifying a fraction of training data in order to reduce the test accuracy. Our framework enables attackers to flexibly adjust the attack's focus on prediction accuracy or fairness and accurately quantify the impact of each candidate point to both accuracy loss and fairness violation, thus producing effective poisoning samples. Experiments on two real datasets demonstrate the effectiveness and efficiency of our framework.
\end{abstract}
\section{Introduction}
Both fair machine learning and adversarial machine learning have received increasing attention in past years. Fair machine learning (FML) aims to learn a function for a target variable  using input features, while ensuring the predicted value be fair with respect to some sensitive attributes based on given fairness criterion. FML models can be categorized into pre-processing, in-processing, and post-processing (see a survey \cite{mehrabi2019survey}). Adversarial machine learning focuses on vulnerabilities in machine learning models and has been extensively studied from perspectives of attack settings and defense strategies (see surveys \cite{yuan2019adversarial,chakraborty2018adversarial}).

There have been a few works on attacking FML models very recently.
\citet{solans2020poisoning} developed a gradient-based poisoning attack to increase demographic disparities among different groups.
\citet{DBLP:journals/corr/abs-2012-08723} also focused on demographic disparity and presented anchoring attack and influence attack.
\citet{chang2020adversarial} focused on attacking FML models with equalized odds. To tackle the challenge of intractable constrained optimization, they developed approximate algorithms for generating poisoning samples. However, how to effectively generate poisoning samples to attack algorithmic fairness still remains challenging due to its difficulty of quantifying impact of each poisoning sample to accuracy loss or fairness violation in the trained FML model.

In this paper, we present a poisoning sample based framework (PFML) for attacking fair machine learning models. The framework enables attackers to adjust their attack's focus on either decreasing prediction accuracy or increasing fairness violation in the trained FML model. Our framework supports a variety of group based fairness notions such as demographic parity and equalized odds. We present three training-time attacks, adversarial sampling, adversarial labeling, and adversarial feature modification.   All of these attacks leave the test data unchanged and instead perturb the training data to affect the learned FML model. In adversarial sampling, the attacker is restricted to select a subset of samples from a candidate attack dataset that has the same underlying distribution of the clean data. Adversarial labeling and adversarial feature modification  can further flip the labels or modify features of selected samples. All three developed attacking methods are online attacks, which are more efficient than those offline poisoning attacks. Our framework enables attackers to flexibly adjust the attack's focus on prediction accuracy or fairness and accurately quantify the impact of each candidate point to both accuracy loss and fairness violation, thus producing effective poisoning samples. Experiments on two real datasets demonstrate the effectiveness and efficiency of our framework.


\section{Background}

\subsection{Fair Machine Learning}
Consider a binary classification task $f_{\theta}: \mathcal{X}\rightarrow \mathcal{Y}$ from an input $x\in \mathcal{X}$ to an output $y \in \mathcal{Y}$. 
Let $l: \Theta \times \mathcal{X} \times \mathcal{Y} \rightarrow \mathbb{R}_{+}$  be a loss function, $\mathcal{D}$ be the training set and each $(x,y) \in \mathcal{D}$ be a data point. The classification model minimizes,  $\mathcal{L}(\theta, \mathcal{D}) = \sum_{(x,y) \in \mathcal{D}}{l(\theta; x, y)}$, the cumulative loss of the model over the training data set $\mathcal{D}$, to obtain the optimal parameters.
Without loss of generality, we assume $\mathcal{X}$  contains one binary sensitive feature $S \in \{0,1\}$.  FML aims to train a model such that its predictions are fair with respect to $S$ based on a given fairness notion, e.g., disparate impact, equal opportunity and equalized odds.

\begin{definition}
\label{def:fml}
A binary classifier $f_{\theta}$ is $\delta$-fair under a fairness notion $\Delta$ if $\Delta(\theta,\mathcal{D}) \leq\delta$, where
$\Delta(\theta,\mathcal{D})$ is referred as the empirical fairness gap of the model and $\delta$ is a user-specified threshold. The model satisfies exact fairness when $\delta =0$.
\end{definition}

\begin{definition}\label{def:Demographic_parity}
We denote demographic parity and equalized odds as $\Delta_{DP}$ and $\Delta_{EO}$, respectively. They are defined as:
\begin{equation}
\label{eq:DP}
\begin{split}
 \Delta_{DP}(\theta,\mathcal{D}) := |\Pr(f_{\theta}(X)=1|S=1) \\  -\Pr(f_{\theta}(X)=1|S=0)|
 \end{split}
 \end{equation}
 \begin{equation}
 \label{eq:EO}
    \begin{split}
        \Delta_{EO}(\theta,\mathcal{D}):=\max_{y\in\{0,1\}}|{Pr}[f_{\theta}(X)\neq y|S=0,Y=y]\\-{Pr}[f_{\theta}(X)\neq y|S=1, Y=y]|
    \end{split}
\end{equation}
\end{definition}

Demographic parity requires that the predicted labels are independent of the protected attribute.  Equalized odds \cite{hardt2016equality}  requires the protected feature $S$ and predicted outcome $\hat{Y}$ are conditionally independent given the true label $Y$.  Equalized opportunity is a weaker notion of equalized odds and requires non-discrimination only within the advantaged outcome group. Our framework naturally covers equalized opportunity. The FML model achieves $\delta$-fairness empirically by minimizing the model's empirical accuracy loss under the fairness constraint:
 \begin{equation}
\begin{aligned}
\hat{\theta} =  \underset{\theta \in \Theta}{\text{arg min}} \frac{1}{|\mathcal{D}|} \mathcal{L}(\theta; \mathcal{D}) \hspace{0.1cm} \text{s.t.} \hspace{0.2cm} C(\theta, \mathcal{D}) =  \Delta\left(\theta,\mathcal{D}\right) - \delta \leq 0
\end{aligned}
\end{equation}

\subsection{Data Poisoning Attack}

Data poisoning attacks \cite{barreno2006can, DBLP:conf/icml/BiggioNL12,mei2015using}  seek to increase the misclassification rate for test data by perturbing the training data to affect the learned model. The perturbation can generally include inserting, modifying or deleting points from the training data  so that the trained classification model can change its decision boundaries and thus yields an adversarial output. The modification can be done by either directly modifying the labels of the training data or manipulating the input features depending on the adversary's capabilities.  In this study, we assume that an attacker can access to the training data during the data preparation process and have the knowledge of the structure and fairness constraint of the classification model.
 We focus on three data poisoning attacks, adversarial sampling, adversarial labeling, and adversarial feature modification, against group-based FML models. In all three attacks, the adversary can select the feature vector of the poisoning data from an attack dataset $\mathcal{D}_k$, which is sampled from the same underlying distribution of the clean dataset $\mathcal{D}_c$, and can control sampling, labeling, or modifying for a fraction of training data in order to reduce the test accuracy.

\begin{algorithm}[t]
	\caption{Online Learning for Generating Poisoning Data}
	\label{alg:poisoning}
	\begin{algorithmic}[1]
		\REQUIRE $\mathcal{D}_c$, $n = |\mathcal{D}_c|$, feasible poisoning set $\mathcal{F}(\mathcal{D}_k)$, number of poisoning data $\epsilon n$,  learning rate $\eta$. \\
		\ENSURE ~~Poisoning dataset $\mathcal{D}_p$. \\
		\STATE Initialize $\theta^0\in\Theta$, $\mathcal{D}_p \leftarrow Null$ \\
		\STATE \textbf{for} $t = 1 : \epsilon n$
		\STATE \hspace{0.2cm}$(x^t,y^t) \leftarrow argmax_{(x,y)\in\mathcal{F}(\mathcal{D}_k)}[ l(\theta^{t-1};x,y)$ \\
        \STATE \hspace{0.2cm} $\mathcal{D}_p\leftarrow \mathcal{D}_p\cup\left\{\left(x^t,y^t\right)\right\}$, $\mathcal{F}(\mathcal{D}_k) \leftarrow \mathcal{F}(\mathcal{D}_k) - \left\{(x,y)\right\} $ \\
		\STATE \hspace{0.2cm} $\theta^t\leftarrow\theta^{t-1}-\eta\frac{\nabla\mathcal{L}\left(\theta^{t-1};\mathcal{D}_c\cup\mathcal{D}_p\right)}{n+t} $
		\STATE \textbf{end for}
	\end{algorithmic}
\end{algorithm}

Algorithm \ref{alg:poisoning} shows the general online gradient descent algorithm for generating poisoning samples. The input parameter $n$ denotes the size of the clean set $\mathcal{D}_c$, $\epsilon$ is the fraction of the size of generated poisoning data over the clean data in the training data set, $\mathcal{F}(\mathcal{D}_k)$ is feasible poisoning set.
Specifically,  $\mathcal{F}(\mathcal{D}_k)$ is the same as $\mathcal{D}_k$ for adversarial sampling. A fraction of data points  $(x,y)\in \mathcal{D}_k$ are changed to $(x,1-y)$ for adversarial labeling, and to $(\tilde{x},y)$ for adversarial feature modification  where $\tilde{x}$ is a modified version of feature vector $x$.
In line 1, it first initializes the model with $\theta^0\in\Theta$. Using the feasible set of poisoning points, the algorithm generates $\epsilon n $ poisoning data points iteratively. In line 3, it selects a data point with the highest impact on the loss function with respect to $\theta^{t-1}$. In line 4, it adds the generated data point to $\mathcal{D}_p$. In line 5, the model parameters $\theta$ are updated to minimize the loss function based on the selected data point  $(x^t,y^t)$.

\section{Data Poisoning Attack on FML}\label{sec: algorithm}
\subsection{Problem Formulation}

We formulate the data poisoning attack on algorithmic fairness as a bi-level optimization problem:
\begin{equation}\label{eq:bi-level_framework}
\begin{aligned}
&\max_{\mathcal{D}_p} \mathbb{E}_{(x, y)}[\alpha \cdot l(\hat{\theta}; x, y) + (1-\alpha) \cdot \gamma \cdot  l_f(\hat{\theta}; x, y)] \\
&\text{where} \hspace{0.2cm}\hat{\theta} =  \underset{\theta \in \Theta}{\text{arg min}} \hspace{0.1cm} \frac{\mathcal{L}(\theta; \mathcal{D}_c\cup \mathcal{D}_p)} {| \mathcal{D}_c\cup \mathcal{D}_p|} \\
& \hspace{1.1cm}\text{s.t.} \hspace{0.2cm}C(\theta, \mathcal{D}_c\cup \mathcal{D}_p) = \Delta\left(\theta, \mathcal{D}_c\cup \mathcal{D}_p \right) - \delta \leq 0
\end{aligned}
\end{equation}
where $\alpha \in [0,1]$ is a hyperparameter that controls the balance of the attack's focus on accuracy and fairness,  $l(\hat{\theta}; x, y)$ is the prediction accuracy loss of the sample $(x, y)$,   $l_f(\hat{\theta}; x, y)$ is the fairness loss, and $\gamma$ is a hyperparameter to have $l_c$ and $l_f$ at the same scale.

We can solve Eq. \ref{eq:bi-level_framework} by optimizing user and attacker's objectives separately. Intuitively, the user (inner optimization) minimizes the classification loss subject to fairness constraint.  The attacker (outer optimization) tries to maximize the joint loss $ \mathbb{E}_{(x, y)}[\alpha \cdot l(\hat{\theta}; x, y) + (1 - \alpha) \cdot \gamma \cdot l_f(\hat{\theta}; x, y)]$ by creating a poisoning set $\mathcal{D}_p$ based on $\hat{\theta}$ obtained by the user to degrade the performance of classifier either from accuracy or fairness aspect. For example, if the value of $\alpha$ approaches to 1, then the attacker tends to degrade more on the accuracy of the model. Note that the loss expectation is taken over the underlying distribution of the clean data.
We approximate the loss function in the outer optimization via the loss on the clean training data and the poisoning data and  have
\begin{equation}
\label{eq:pfml-outer}
\max_{\mathcal{D}_p} [\alpha \cdot \mathcal{L}(\hat{\theta}; \mathcal{D}_c\cup \mathcal{D}_p) + (1-\alpha) \cdot \gamma \cdot  \Delta(\hat{\theta}; \mathcal{D}_c\cup \mathcal{D}_p)]
\end{equation}

\begin{figure}[htb]
    \centering
    \includegraphics[width=0.4\textwidth]{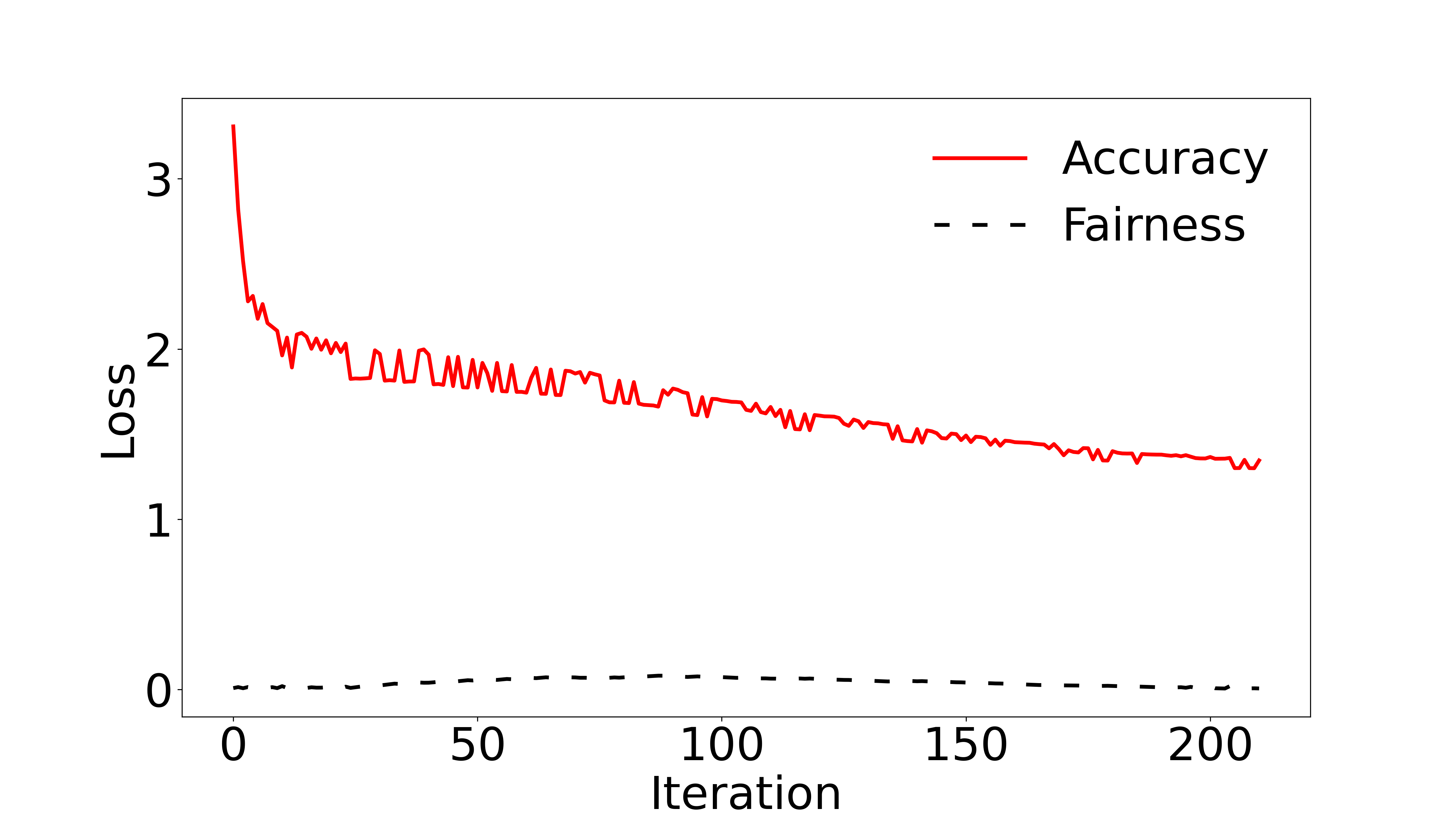}
    \caption{Accuracy and fairness loss (in terms of equalized odds) with different iterations of PFML-AS ($\alpha=0.8$) on COMPAS.}
    \label{fig:loss}
\end{figure}

The accuracy loss $\mathcal{L}$ and fairness loss $\Delta$ may be at different scales due to the use of different loss functions and data distribution.  Figure \ref{fig:loss} shows the curves of accuracy loss and fairness loss of equalized odds  when we increase the generated poisoning samples from 1 to 211 on COMPAS dataset (see experiment section for the detailed experimental setting). This shows the importance of introducing hyperparameter $\gamma$ to have accuracy loss and fairness loss at the same scale.

The user tries to achieve optimal and fair $\hat{\theta}$ under the poisoning set $\mathcal{D}_p$.
As the constrained optimization is intractable, we further transform the inner optimization to its dual form as the following:
\begin{equation}
\label{eq:pfml-inner}
\hat{\theta} = \min_{\theta \in \Theta} \left(\frac{1}{n+t}\mathcal{L}(\theta; \mathcal{D}_c\cup \mathcal{D}_p) + \lambda  \Delta(\theta, \mathcal{D}_c\cup \mathcal{D}_p)\right)
\end{equation}
where $\lambda$ is the Lagrange multiplier and $t$ is the current size of poisoning samples $\mathcal{D}_p$.
By Eq. \ref{eq:pfml-outer} and Eq. \ref{eq:pfml-inner}, we effectively capture the contribution of each poisoning point $(x,y)$ to both accuracy loss and fairness gap.

\subsection{Convex Relaxation of Fairness Constraint}

The dual optimization problem in  Eq. \ref{eq:pfml-inner} involves the calculation of $\Delta(\theta, \mathcal{D}_c\cup \mathcal{D}_p)$ over the current  $\mathcal{D}_c\cup \mathcal{D}_p$. However, fairness notions such as demographic parity and equalized odds are non-convex. We adopt simplifications proposed by \cite{zafar2017fairness} for demographic parity  and  \cite{DBLP:conf/nips/DoniniOBSP18} for equalized odds to reach convex relaxations of fairness constraints.

Demographic parity can be approximated by the decision boundary fairness. The decision boundary fairness over $\mathcal{D}_c \cup \mathcal{D}_p$ is defined as the covariance between the sensitive attribute and the signed distance from the non-sensitive attribute vector to the decision boundary. It can be written as:
\begin{equation}\label{eq:fairness_convariance}
   C(\theta, \mathcal{D}_c\cup \mathcal{D}_p) = \dfrac{1}{n + t}\sum_{i = 1}^{n + t}(s_i- \bar{s})d_{\theta({x}_i)}
\end{equation}
where $t$ is the size of the current poisoning samples $\mathcal{D}_p$, $s_i$ is the value of the sensitive attribute of the sample ${x}_i$, $d_{\theta({x}_i)} = \theta^T x_i$ is the distance to the decision boundary of the classifier $f_{\theta}$, $\bar{s}$ is the mean value of the sensitive attribute over $\mathcal{D}_c \cup \mathcal{D}_p$.
We require that $|C(\theta, \mathcal{D}_c\cup \mathcal{D}_p)| \leq \tau$ to achieve fairness.

We adopt the fairness definition for equalized odds  by balancing the risk among two sensitive groups. Let the linear loss be $\mathcal{L}_l$ (e.g., $\mathcal{L}_l = 0.5({1-f_{\theta}(x)})$ for SVM model) and  denote $\mathcal{D} = \mathcal{D}_c\cup \mathcal{D}_p$. We can write down the convex
relaxation for the fairness gap of equalized odds as the following:
\begin{equation}
    C(\theta, \mathcal{D}) = \dfrac{1}{2} \sum_{y=0,1}|R^{y, s=0}(\theta, \mathcal{D})- R^{y, s=1}(\theta, \mathcal{D})|
\end{equation}
where $R^{y, s}(\theta, \mathcal{D}) = \dfrac{1}{n^{y,s}}\sum_{(x, y) \in \mathcal{D}_{y, s}}\mathcal{L}_l(x, y; \theta)$. $\mathcal{D}_{y, s}$ is the dataset of points with group $s$ and label $y$ and $n^{y,s}$ is the size of $\mathcal{D}_{y, s}$.

\begin{algorithm}[htb]
	\caption{Poisoning Attack on Fair Machine Learning (PFML) }
	\label{alg:PFML}
	\begin{algorithmic}[1]
		\REQUIRE $\mathcal{D}_c$, $n = |\mathcal{D}_c|$, feasible poisoning set $\mathcal{F}(\mathcal{D}_k)$, number of poisoning data $\epsilon n$, penalty parameter (Lagranger multiplier) $\lambda$, learning rate $\eta$, scaling factor $\gamma$, balance ratio $\alpha$, fairness notion $\Delta$. \\
		\ENSURE ~~Poisoning dataset $\mathcal{D}_p$. \\
		\STATE Initialize $\theta^0\in\Theta$
		\STATE \textbf{for} $i = 1 : I$
		\STATE \hspace{0.5cm}\scalebox{0.95}{$\theta^i\leftarrow\theta^{i-1}-\eta\left(\frac{\nabla\mathcal{L}\left(\theta^{i-1};\mathcal{D}_c    \right)}{n}+
        \nabla\left[\lambda
        \Delta\left(\theta^{i-1},\mathcal{D}_c\right)\right]\right)$} \\
		\STATE \textbf{end for}
		\STATE $\theta^0 \leftarrow \theta^{I}$, $\mathcal{D}_p \leftarrow Null$
		\STATE \textbf{for} $t = 1 : \epsilon n$
		\STATE \hspace{0.2cm}$(x^t,y^t) \leftarrow argmax_{(x,y)\in\mathcal{F}(\mathcal{D}_k)}[ \alpha\cdot l(\theta^{t-1};x,y) + $\\
		\STATE \hspace{2.5cm}\scalebox{0.9}{$(1-\alpha)\cdot\gamma\cdot\Delta\left(\theta^{t-1},\mathcal{D}_c\cup\mathcal{D}_p\cup\left\{(x,y)\right\}\right)]$}\\
        \STATE \hspace{0.2cm} $\mathcal{D}_p\leftarrow \mathcal{D}_p\cup\left\{\left(x^t,y^t\right)\right\}$, $\mathcal{F}(\mathcal{D}_k) \leftarrow \mathcal{F}(\mathcal{D}_k) - \left\{(x,y)\right\} $
		\STATE \hspace{0.2cm} \scalebox{0.85}{$\theta^t\leftarrow\theta^{t-1}-\eta\left(\frac{\nabla\mathcal{L}\left(\theta^{t-1};\mathcal{D}_c\cup\mathcal{D}_p    \right)}{n+t}+
        \nabla\left[\lambda
        \Delta\left(\theta^{t-1},\mathcal{D}_c\cup\mathcal{D}_p\right)\right]\right)$}\\
		\STATE \textbf{end for}
	\end{algorithmic}
\end{algorithm}

\begin{table*}[htb]
	\caption{Test accuracy and fairness gap of fair reduction \cite{agarwal2018reductions} and post-processing~\cite{hardt2016equality} with {\bf equalized odds} under PFML and baselines (COMPAS).}
	\centering
	\scalebox{0.78}{
	\begin{tabular}{|c|c|c|c|c|c||c|c|c|c|c|}
		\hline
		 & \multicolumn{5}{|c||}{Accuracy} & \multicolumn{5}{|c|}{Fairness} \\ \hline
		Methods & \multicolumn{4}{|c|}{Fair Reduction } & \multicolumn{1}{|c||}{Post-processing}  & \multicolumn{4}{|c|}{Fair Reduction} & \multicolumn{1}{|c|}{Post-processing} \\ \hline
		
		\hline
		\ & $\delta = 0.12$ & $\delta = 0.1$ & $\delta = 0.07$ & $\delta = 0.05$ & $\delta = 0$& $\delta = 0.12$ & $\delta = 0.1$ & $\delta = 0.07$ & $\delta = 0.05$& $\delta = 0$\\
		\hline
		Benign & 0.950&	0.949&	0.949&	0.948& 0.877& 0.108&	0.103&	0.086&	0.082 & 0.095 \\\hline
		RS &  0.936&	0.930&	0.919&	0.912&	0.839 &  0.101&	0.105&	0.104&	0.103&	0.081\\
		\hline
		LF & 0.935&	0.931&	0.919&	0.911&	0.839 & 0.062&	0.066&	0.072&	0.080&	0.109\\
		\hline
		HE &  0.915&	0.908&	0.899&	0.891&	0.829 &  0.076&	0.082&	0.100&	0.109&	0.131\\
		\hline
	    INFL   & 0.850 &	0.848&	0.845&	0.841&	 0.653&  0.089&	0.081&	0.078&	0.081&	0.054\\	\hline
	    KKT    & 0.890 &	0.891&	0.891&	0.886&	 0.701&  0.136&	0.137&	0.137&	0.142&	0.096\\	\hline
		min-max &  0.891&	0.887&	0.878&	0.874&	 0.678&  0.096&	0.125&	0.089&	0.075&	0.082\\	\hline
		AS  &  0.830&	0.824&	0.816&	0.810&	0.740 &  0.051&	0.069&	0.111&	0.143&	0.156 \\
		\hline
		AF &  0.823&	0.817&	0.808&	0.803&	0.740 &  0.046&	0.059&	0.100&	0.130&	0.136\\
		\hline
		\hline
	PFML-AS, $\alpha = 0$ & 0.853&    0.847&  0.833&  0.802&  0.753 & 0.126&    0.148&  0.164&  0.185&  0.190 \\\hline
	PFML-AS, $\alpha = 0.2$ & 0.843&    0.837&  0.820&  0.792&  0.728 & 0.112&    0.124&  0.138&  0.127&  0.188 \\\hline
	PFML-AS, $\alpha = 0.5$ &  0.824&    0.820&  0.814&  0.809&  0.705 & 0.110&    0.118&  0.130&  0.142&  0.143 \\\hline
	PFML-AS, $\alpha = 0.8$ & 0.820&    0.816&  0.809&  0.800&  0.715 & 0.101&    0.105&  0.116&  0.120&  0.099  \\\hline
	PFML-AS, $\alpha = 1.0$ & 0.811&    0.807&  0.800&  0.796&  0.724 & 0.083&    0.071&  0.061&  0.061&  0.074\\\hline \hline
	PFML-AF, $\alpha = 0$ & 0.847&	0.841&	0.832&	0.805&	0.752 & 0.120&	0.144&	0.172&	0.184&	0.193\\\hline
	PFML-AF, $\alpha = 0.2$ & 0.843&	0.838&	0.817&	0.792&	0.728 & 0.107&	0.117&	0.125&	0.126&	0.186 \\\hline
	PFML-AF, $\alpha = 0.5$ & 0.818&	0.814&	0.808&	0.804&	0.711 & 0.101&	0.110&	0.126&	0.139&	0.136\\\hline
	PFML-AF, $\alpha = 0.8$ & 0.804&	0.797&	0.791&	0.786&	0.714 & 0.093&	0.090&	0.097&	0.107&	0.090\\\hline
	PFML-AF, $\alpha = 1.0$ & 0.803&	0.797&	0.794&	0.788&	0.722 & 0.088&	0.068&	0.043&	0.039&	0.097\\\hline \hline
	PFML-AM, $\alpha = 0$   & 0.908	&0.906	&0.904	&0.897	&0.764 & 0.195	&0.200	&0.215	&0.207	&0.198\\\hline
	PFML-AM, $\alpha = 0.2$ & 0.811	&0.805	&0.798	&0.794	&0.731 & 0.102	&0.086	&0.076	&0.076	&0.153\\\hline
	PFML-AM, $\alpha = 0.5$ & 0.793	&0.788	&0.780	&0.775	&0.688 & 0.079	&0.059	&0.071	&0.080	&0.124\\\hline
	PFML-AM, $\alpha = 0.8$ & 0.791	&0.789	&0.782	&0.773	&0.696 & 0.082	&0.055	&0.053	&0.077	&0.096\\\hline
	PFML-AM, $\alpha = 1.0$ & 0.828	&0.823	&0.817	&0.813	&0.696 & 0.063	&0.045	&0.055	&0.073	&0.076\\\hline
	\end{tabular}}\label{tab:compas}
\end{table*}

\begin{table*}[htb]
	\caption{Test accuracy and fairness gap of fair reduction  and  post-processing with {\bf demographic parity} (COMPAS).}
	\centering
	\scalebox{0.78}{
	\begin{tabular}{|c|c|c|c|c|c||c|c|c|c|c|}
		\hline
		 & \multicolumn{5}{|c||}{Accuracy} & \multicolumn{5}{|c|}{Fairness} \\ \hline
		Methods & \multicolumn{4}{|c|}{Fair Reduction } & \multicolumn{1}{|c||}{Post-processing}  & \multicolumn{4}{|c|}{Fair Reduction} & \multicolumn{1}{|c|}{Post-processing} \\ \hline
		
		\hline
		 & $\delta = 0.12$ & $\delta = 0.1$ & $\delta = 0.07$ & $\delta = 0.05$ & $\delta = 0$ & $\delta = 0.12$ & $\delta = 0.1$ & $\delta = 0.07$ & $\delta = 0.05$ & $\delta = 0$ \\
		\hline
	Benign & 0.887&	0.867&	0.803&	0.768& 0.859& 0.175&	0.169&	0.107&	0.095 & 0.046 \\\hline
	RS & 0.882&	0.839&	0.813&	0.767& 0.867&0.187&	0.155&	0.130&	0.076 & 0.023 \\\hline
	LF & 0.890&	0.852&	0.814&	0.775& 0.868& 0.194&	0.166&	0.138&	0.099 & 0.021 \\\hline
	HE & 0.901&	0.859&	0.808&	0.766& 0.840&0.205&	0.181&	0.135&	0.098 & 0.036 \\\hline
	INFL & 0.879&  0.855  & 0.774 & 0.748 & 0.784 & 0.200 & 0.186 & 0.097 & 0.108& 0.015  \\\hline
	KKT & 0.884&  0.875  & 0.788 & 0.768 & 0.817 & 0.221 & 0.214 & 0.127 & 0.136 & 0.016 \\\hline
	min-max & 0.870&    0.870&  0.843&  0.818&  0.810&  0.201&  0.204&  0.182& 0.167& 0.036\\\hline\hline
	PFML-AS, $\alpha = 0$   & 0.853&	0.829&	0.771&	0.750&	0.824& 0.195&	0.168&	0.109&	0.099&	0.041  \\\hline
	PFML-AS, $\alpha = 0.2$ & 0.847&	0.819&	0.766&	0.736&	0.798& 0.189&	0.171&	0.106&	0.100&	0.039\\\hline
	PFML-AS, $\alpha = 0.5$ & 0.844&	0.812&	0.763&	0.731&	0.795& 0.182&	0.167&	0.101&	0.092&	0.038\\\hline
	PFML-AS, $\alpha = 0.8$ & 0.845&	0.811&	0.758&	0.731&	0.791& 0.175&	0.166&	0.096&	0.094&	0.036\\\hline
	PFML-AS, $\alpha = 1.0$ & 0.829&	0.816&	0.757&	0.722&	0.790& 0.171&	0.151&	0.083&	0.075&	0.032\\\hline\hline
	PFML-AF, $\alpha = 0$   & 0.848&	0.822&	0.786&	0.761&	0.822& 0.192&	0.185&	0.098&	0.080&	0.057  \\\hline
	PFML-AF, $\alpha = 0.2$ & 0.841&	0.805&	0.766&	0.742&	0.806& 0.188&	0.163&	0.095&	0.086&	0.056\\\hline
	PFML-AF, $\alpha = 0.5$ & 0.842&	0.809&	0.762&	0.733&	0.801& 0.174&	0.136&	0.087&	0.086&	0.036\\\hline
	PFML-AF, $\alpha = 0.8$ & 0.838&	0.803&	0.755&	0.729&	0.798& 0.167&	0.134&	0.086&	0.079&	0.027\\\hline
	PFML-AF, $\alpha = 1.0$ & 0.831&	0.808&	0.752&	0.721&	0.793& 0.160&	0.132&	0.082&	0.069&	0.032\\\hline\hline
	PFML-AM, $\alpha = 0$   & 0.883&	0.853&	0.816&	0.791&	0.833& 0.246&	0.219&	0.183&	0.159&	0.031  \\\hline
	PFML-AM, $\alpha = 0.2$ & 0.840&	0.820&	0.762&	0.730&	0.814& 0.218&	0.208&	0.138&	0.128&	0.038\\\hline
	PFML-AM, $\alpha = 0.5$ & 0.838&	0.802&	0.757&	0.733&	0.793& 0.212&	0.170&	0.120&	0.114&	0.030\\\hline
	PFML-AM, $\alpha = 0.8$ & 0.826&	0.800&	0.758&	0.720&	0.787& 0.193&	0.147&	0.115&	0.065&	0.031\\\hline
	PFML-AM, $\alpha = 1.0$ & 0.853&	0.805&	0.767&	0.726&	0.815& 0.184&	0.138&	0.105&	0.060&	0.029\\\hline
	\end{tabular}}\label{tab:compas_rd}
\end{table*}

\subsection{Attack Algorithm}\label{sec:algorithm}

Algorithm \ref{alg:PFML} shows pseudo code of our  poisoning attack framework on fair machine learning (PFML). Our three algorithms are denoted as PFML-AS for adversarial sampling, PFML-AF for adversarial flipping, and PFML-AM for adversarial feature modification. In each algorithm, we can adjust the attack's focus on prediction accuracy or fairness by choosing different $\alpha$ values. For example, when $\alpha = 1$ ($0$),  the attack's focus is purely on accuracy (fairness) and  when $\alpha = 0.5$, the focus is on the combination of fairness and accuracy. In line 2 - 4, we first train FML model on the clean data $\mathcal{D}_c$ and use the fitted parameter $\theta^I$ to start generating poisoning samples.  We then execute the loop of line 6 - 9 to iteratively generate $\epsilon n$ poisoning samples. In line 7, when generating the data point $(x^t,y^t)$ with highest impact on a weighted sum of the accuracy loss and the fairness violation with respect to $\theta^{t-1}$, we add both the previously generated data points in $\mathcal{D}_p$ and the data point $(x^t,y^t)$ to $\mathcal{D}_c$. As a result, we can measure the incremental contribution of that data point to the fairness gap  $\Delta\left(\theta^{t-1},\mathcal{D}_c\cup\mathcal{D}_p\cup\left\{(x,y)\right\}\right)$. Note that in this step, the accuracy loss can be simply calculated over each point $(x,y) \in \mathcal{F}(\mathcal{D}_k)$ as the accuracy loss of existing data points from $\mathcal{D}_c \cup \mathcal{D}_p$ is unchanged.
In line 8, we add the chosen poisoning point $(x^t,y^t)$ to $\mathcal{D}_p$ and also remove it from the feasible poisoning set.
In line 9, when updating the model parameters $\theta$, we minimize the penalized loss function over $\mathcal{D}_c$ and $\mathcal{D}_p$.
We see the execution time is mostly spent on line 6 - 11. In fact, line 9 and line 10 only involve one time operation. The time complexity of line 8 is $\mathcal{O}(m)$, where $m$ is the size of feasible poisoning set $\mathcal{F}(\mathcal{D}_k)$. Therefore, the time complexity of the loop from line 6 - 11 is $\mathcal{O}(\epsilon nm)$.

{\noindent \bf Remarks.}
\citet{chang2020adversarial} presented an online gradient descent algorithm that generates poisoning data points for fair machine learning model with equalized odds. As the fairness gap is not an additive function of the training data points, they used $\mathcal{D}_c\cup\left\{(x^t,y^t)^{\epsilon n}\right\}$ (denoted as $\mathcal{D}_t$)  to measure  the contribution of that data point to the fairness gap where $\mathcal{D}_t$ is equivalent to adding $\epsilon n$ copies of  $(x^t,y^t)$ to $\mathcal{D}_c$. The weighted loss function used for selecting poisoning samples is shown as $\left[\epsilon\cdot l\left(\theta^{t-1};x,y\right)+\lambda\cdot\Delta\left(\theta^{t-1},\mathcal{D}_t\right)\right]$. The algorithm then updates the model parameters $\theta$ via the gradient descent, i.e., $\theta^t\leftarrow\theta^{t-1}-\eta(\frac{\nabla\mathcal{L}\left(\theta^{t-1};\mathcal{D}_c\right)}{n}+\nabla\left[\epsilon\cdot l\left(\theta^{t-1};x^t,y^t\right)+\lambda\cdot\Delta\left(\theta^{t-1},\mathcal{D}_t\right)\right])$. However, both the use of $\mathcal{D}_c\cup\left\{(x,y)^{\epsilon n}\right\}$ to quantify the $(x^t,y^t)$'s contribution to the fairness gap and the use of parameters ($\epsilon$ and $\lambda$) to define the weighted loss are heuristic, thus hard to produce effective poisoning samples on algorithmic fairness. Moreover, different from \citep{chang2020adversarial} that covers only a single fairness notion (i.e., equalized odds) and two attacks (adversarial sampling and adversarial label flipping), our paper presents a general framework with algorithms for three group based fairness notions and a new important adversarial feature modification attack.  In our evaluation, we compare our methods with \citet{chang2020adversarial} and three new state-of-the-art baselines (influence attack, KKT, and min-max attack) from  \cite{koh2018stronger}.

\section{Experiments}
\label{sec:exp}

{\bf \noindent Datasets.} We conduct our experiments on COMPAS \cite{compass} and Adult \cite{adult} which are two benchmark datasets for FML community.
We use race (only black/white) as the sensitive attribute for COMPAS and gender as sensitive attribute for Adult. After preprocessing, COMPAS has 5278 data points and 11 features, while Adult has 48842 data points and 14 features.
For each dataset, we first train a SVM model on the entire dataset. For the 60\% data with the smallest loss, we randomly split them into clean dataset $\mathcal{D}_c$, attack candidate dataset $\mathcal{D}_k$, and test dataset $\mathcal{D}_{test}$ with ratio 4:1:1. The rest 40\% data is treated as hard examples and added into $\mathcal{D}_k$. For COMPAS, $\mathcal{D}_c$ contains 2111 samples and $\mathcal{D}_{test}$ has 528 samples. $\mathcal{D}_{k}$ has 2639 samples including 2112 hard examples.
For adversarial label flipping, we randomly flip the label of 15\% data from $\mathcal{D}_{k}$ to build the feasible poisoning candidate set $\mathcal{F}(\mathcal{D}_k)$. For adversarial feature modification, we randomly flip one binary feature of each data point from $\mathcal{D}_k$ and include them to  $\mathcal{F}(\mathcal{D}_k)$.
Following the similar pre-processing strategy, $\mathcal{D}_c$,  $\mathcal{D}_{test}$, and $\mathcal{D}_k$ of Adult contain 15385 samples, 6594 samples, and 26863 samples, respectively.
We report detailed results of  COMPAS in the majority of this experiment section and only show the summarized results of Adult in Figure \ref{fig:adult} at the end of this experiment section.

{\bf \noindent Baselines.}
We consider the following baselines: (a) Random Sampling (RS): attacker randomly selects data samples from $\mathcal{D}_k$; (b) Label Flipping (LF): attacker randomly selects data samples from $\mathcal{D}_k$ and flips their labels; (c) Hard Examples (HE): attacker randomly selects data samples from hard examples set; (d) influence attack (INFL), (e) KKT attack, (f) min-max attack, (g) adversarial sampling (AS),  and (h) adversarial flipping (AF). Attacks (d)-(f) are stronger data poisoning attacks breaking
data sanitization defenses  and all control both the label $y$ and input features $x$ of the poisoned points  \cite{koh2018stronger}. Attacks (g) and (h) are designed for attacking FML from \cite{chang2020adversarial}. As attacks (g) and (h) are only designed for equalized odds, we exclude them from baselines when reporting comparisons based on  demographic parity.

{\bf \noindent Fair Classification Models.} We use SVM as the classification model and choose fair reduction \cite{agarwal2018reductions} and post-processing \cite{hardt2016equality} as FML under attack.  Post-processing  adjusts an unconstrained trained model to remove discrimination based on fairness notions such as demographic parity and equalized odds. After adjustment, the unconstrained model behaves like a randomized classifier that assigns each data point a probability  conditional on its protected attribute and predicted label. These probabilities are calculated by a linear program to minimize the expected loss.
Fair reduction is an advanced in-processing approach that reduces fair classification to a sequence of cost-sensitive classification and achieves better accuracy-fairness tradeoff than previous FML models. Hence, we do not report results from other in-processing FML models.

{\bf \noindent Hyperparameters.} In our default setting, we choose the number of pretrain steps with $\mathcal{D}_c$ as 2000,  learning rate $lr$ as 0.001, penalty parameter $\lambda$ as 5, and $\epsilon$ as 0.1. The scaling factor $\gamma$ is calculated as the ratio of accuracy loss and fairness loss over $\mathcal{D}_c$.

{\bf \noindent Metrics.} We run our attacks, PFML-AS, PFML-AF and PFML-AM,  each with five $\alpha$ values,  and baseline attacks to generate the poisoning data $\mathcal{D}_p$ and then train fair reduction (with four $\delta$ values as fairness threshold) and post-processing models with $\mathcal{D}_c \cup \mathcal{D}_p$.  Finally we run the trained FML models on the test data $\mathcal{D}_{test}$ and report the test accuracy and fairness gap. For each experiment,  we report the average value of five runs.
We skip reporting their standard deviation and instead we summarize comparisons based on t-test. We include in the supplemental file all source code and setting details (e.g., seeds and hyperparameters) for reproducibility.

\subsection{Evaluation of PFML  with Equalized Odds}

Table \ref{tab:compas} shows the comparison of our PFML attacks under different $\alpha$ with other baseline models in terms of both accuracy and fairness on two FML models (fair reduction and post-processing) trained with equalized odds under different fairness threshold values of $\delta$.

First, the accuracy of FML model under all three PFML attacks (PFML-AS, PFML-AF and PFML-AM) is significantly lower than the benign case.   For each fixed $\delta$, both the accuracy value and fairness gap of FML under PFML attacks decrease when we increase $\alpha$. Recall that larger $\alpha$ indicates that PFML attacks more on accuracy and smaller $\alpha$ indicates more attack's focus on  fairness. Note that larger fairness gap caused by smaller $\alpha$ indicates higher model unfairness.
Taking PFML-AS  as an example, the accuracy of fair reduction with $\delta = 0.12$ is 0.853 and the fairness gap is 0.126 when $\alpha=0$; the accuracy is 0.811 and the fairness gap is 0.083 when $\alpha=1$.
This result demonstrates that controlling $\alpha$  is flexible and effective for attackers to tune attack target on either prediction accuracy or fairness. Second,  PFML-AF and PFML-AM outperform PFML-AS in terms of attacking performance from both accuracy and fairness perspectives, which shows modifying input features or flipping labels is more powerful than adversarial sampling.  Third, compared to RS, LF and HE, our PFML attacks can reduce more accuracy or incur more unfairness  with the same $\delta$ for both fair reduction and post-processing. Taking PFML-AS with $\delta = 0.12$ and $\alpha = 0$ as an example, the accuracy is 0.811, which is 0.125, 0.124, and 0.104 lower than that of RS, LF and HE, respectively.
Compared to  previous FML attacks (AS, AF) \cite{chang2020adversarial} and sanitization attacks (INFL, KKT, min-max) \cite{koh2018stronger},  our PFML attacks achieve better attack performance in terms of accuracy (fairness) with large (small) $\alpha$ values, which is consistent with our expectation.

We also notice, for each fixed $\alpha$, the accuracy of fair reduction  under our  PFML attacks decreases when we decrease $\delta$.  The fairness gap of fair reduction under PFML-AS (PFML-AF) attack increases when we decrease $\delta$, which indicates the fair reduction model is less robust or more vulnerable when stricter fairness constraint is enforced. However, the fairness gap of fair reduction under PFML-AM attack instead decreases along the decrease of $\delta$. Theoretical analysis is needed to understand the robustness of fair reduction approach with equalized odds under different poisoning attacks.

\subsection{Evaluation of PFML with Demographic Parity}

Table \ref{tab:compas_rd} shows the comparison results of adversarial fair machine learning with demographic parity. Note that we do not compare with online FML attacks (AS, AF) as they do not support demographic parity. Generally we see similar patterns as equalized odds shown in Table \ref{tab:compas_rd}.
For each fixed $\delta$, both the accuracy and fairness gap of FML models under all three PFML attacks decrease when we increase $\alpha$. This is because smaller $\alpha$ means more attack's focus on fairness.
Compared to RS, LF and HE, our PFML attacks can reduce more model accuracy of FML with the same $\delta$ for both fair reduction and post-processing. Compared to INFL, KKT and min-max attacks,  PFML attacks achieve better attack performance in terms of accuracy drop (fairness gap) of FML models when we set large (small) $\alpha$ values.

For each fixed $\alpha$, the accuracy of fair reduction under PFML attacks decreases when we decrease $\delta$. This pattern is similar as equalized odds. However, the fairness gap of fair reduction has a clear decreasing trend when $\delta$ decreases, which is different from equalized odds. This result actually indicates the fair reduction model with stricter fairness requirement (small $\delta$) is less vulnerable under poisoning attacks.

\subsection{Sensitivity Analysis of Hyperparameters}

In this section, we evaluate the sensitivity of PFML attacks under different hyperparameters.
Table \ref{tab:epsilon} shows the accuracy, fairness gap and execution time for COMPAS with equalized odds when we change the size of poisoning samples $\epsilon$ against fair reduction. In all experiments, we fix $\delta = 0.07$  and $\alpha=0.8$. In general, with increasing $\epsilon$, the accuracy of fair reduction drops while its fairness gap increases when fair reduction is under each of our PMFL attacks. Note that larger $\epsilon$ corresponds to injecting more poisoning data points into the training data, thus causing more accuracy drop and unfairness of the trained FML model.

\begin{table}[htb]
    \centering
      \caption{Effects of ratio $\epsilon$ for fairness reduction with equalized odds (COMPAS).}
    \scalebox{0.7}{\begin{tabular}{|c|c|c|c|c|c|}
    \hline
         Dataset &   & $\epsilon = 0.025$ & $\epsilon = 0.05$ & $\epsilon = 0.1$ &$\epsilon = 0.15$  \\
        \hline
        \hline
        \multirow{6}*{Accuracy}
        & INFL   & 0.891 & 0.857 & 0.845 & 0.820\\  \cline{2-6}
        & KKT  &  0.912 & 0.899 & 0.891 & 0.884\\
        \cline{2-6}
        & min-max  & 0.918 & 0.902 & 0.878 & 0.850\\
        \cline{2-6}
        & PFML-AS & 0.882 & 0.821 & 0.809 &0.799\\
        \cline{2-6}
        & PFML-AF &	0.867&	0.824&	0.791&	0.794 \\
        \cline{2-6}
        & PFML-AM &	0.891&	0.839&	0.782&	0.777 \\ \hline \hline

        \multirow{6}*{Fairness Gap}
        & INFL   & 0.068 & 0.054 & 0.078 & 0.063\\  \cline{2-6}
        & KKT   & 0.086 & 0.102 & 0.137 & 0.158\\
        \cline{2-6}
        & min-max  &  0.083 & 0.108 & 0.089 & 0.215 \\
        \cline{2-6}
        & PFML-AS   & 0.086 & 0.082 &0.116 & 0.134\\
        \cline{2-6}
        & PFML-AF &	0.089&	0.081&	0.097&	0.142 \\
        \cline{2-6}
        & PFML-AM &	0.089&	0.092&	0.077&	0.107 \\ \hline \hline

        \multirow{6}*{Exec. Time (s)}
        & INFL  & 497.1 & 915.7 & 1569.1 & 2009.5 \\  \cline{2-6}
        & KKT  & 1633.6  & 2903.3 & 5503.3 & 8400.0\\
        \cline{2-6}
        & min-max   & 337.9 & 597.1 & 1137.6 & 1714.6 \\
        \cline{2-6}
        & PFML-AS  &	4.8 &	6.1 &	8.8 &	12.1\\
        \cline{2-6}
        & PFML-AF  &	5.1 &	6.7 &	9.9 &	13.9 \\
        \cline{2-6}
        & PFML-AM  &	6.5&	10.9&	16.5&	22.5 \\ \hline

    \end{tabular}}
    \label{tab:epsilon}
\end{table}

We also compare with baseline attack models (INFL, KKT, and min-max). We can see with the same $\epsilon$ our PFML attacks can degrade the model accuracy more than the baselines, and  cause similar or higher level of model unfairness than the baselines in most scenarios.
We also report the execution time in Table \ref{tab:epsilon} and we can see the exeuction time of our PFML attacks increases linearly with increasing $\epsilon$, which is consistent with our time complexity analysis in Algorithm \ref{alg:PFML}.   Compared to the baseline models, our PFML attacks takes two or three orders of magnitude less time to generate poisoning samples than baselines.

\begin{table}[htb]
    \centering
      \caption{Effects of penalty parameter $\lambda$  on PFML-AS for post-processing with equalized odds (COMPAS).}
    \scalebox{0.75}{\begin{tabular}{|c|c|c|c|c|c|c|}
    \hline
         & $\lambda = 1$ & $\lambda = 5$ & $\lambda = 10$ & $\lambda = 15$ & $\lambda = 50$ & $\lambda = 150$  \\
        \hline
        \hline
        Accuracy  & 0.709&	0.715&	0.718&	0.720&	0.726&  0.723\\
        \hline
        Fairness Gap  & 0.094&	0.099&	0.122&	0.118&	0.125&  0.156\\
        \hline
    \end{tabular}}
    \label{tab:lambda}
\end{table}

Table \ref{tab:lambda} shows the accuracy and fairness gap with equalized odds when we use PFML-AS ($\alpha=0.8$) to attack post-processing FML \cite{hardt2016equality} under different $\lambda$ values.  The post-processing approach has strict fairness constraint $\delta =0$. We can see with larger $\lambda$, the PFML attack  focuses more on attacking  fairness, which leads to larger fairness gap and smaller accuracy drop of the FML model.

\begin{table}[htb]
    \centering
      \caption{Significance testing of p-value for PFML on COMPAS (EO: equalized odss; DP: demographic parity).}
    \scalebox{0.7}{\begin{tabular}{|c|c|c|c|c|}
    \hline
        &   & PFML-AS & PFML-AF & PFML-AM  \\
        \hline
        \hline
        \multirow{6}*{accuracy, DP and $\alpha = 1$}
        & RS & 0.00016 & 0.00009 & 0.00008 \\
        \cline{2-5}
        & LF & 0.00003 & 0.00001 & 0.00001 \\
        \cline{2-5}
        & HE & 0.00019 & 0.00011 &0.00009  \\
        \cline{2-5}
        & INFL   & 0.07089  & 0.05140 &0.02517 \\
        \cline{2-5}
        & KKT  & 0.00128 &0.00077  & 0.00050\\
        \cline{2-5}
        & min-max  & 0.00004 &0.00002  & 0.00002\\
        \cline{2-5}
         \hline
       \multirow{8}*{fairness, EO and $\alpha = 0$}
        & RS & 0.00001 & 0.00001  & 0.00000 \\
        \cline{2-5}
        & LF & 0.00000 & 0.00000 & 0.00000 \\
        \cline{2-5}
        & HE & 0.00003 & 0.00004 & 0.00000 \\
        \cline{2-5}
        & INFL   & 0.00000 & 0.00000 & 0.00000\\
        \cline{2-5}
        & KKT  & 0.00472 & 0.00502 & 0.00001\\
        \cline{2-5}
        & min-max  & 0.00002 & 0.00002 & 0.00000\\
        \cline{2-5}
         & AS  & 0.00131  & 0.00136 & 0.00001\\
        \cline{2-5}
        & AF  & 0.00006  & 0.00006 & 0.00000\\
        \cline{2-5}
         \hline
    \end{tabular}}
    \label{tab:p-value}
\end{table}

\subsection{Significance Testing}
\label{sec:t-test}

For each experiment, we have run our methods and other baseline models five times as shown in all our tables.
We apply independent two-sample t-test to compare each of three PFML models (with a given fairness notion and $\alpha$) with each of baseline models in terms of accuracy reduction and fairness respectively.  The t-test results show our PFML attacks significantly outperform baselines from both accuracy and fairness perspectives.
We  report p-values of two cases in Table \ref{tab:p-value}: (C1) demographic parity and $\alpha = 1$ (shown in the first block),  and (C2) equalized odds and $\alpha = 0$ (shown in the second block). C1 focuses on accuracy whereas C2 focuses on fairness. Each p-value in Table \ref{tab:p-value} shows the comparison of one of our PFML attacks against one baseline. We can see 39 out of 42 p-values are less than 0.01 and the left three are still less than 0.1, which demonstrates statistical significance of our PFML methods over baselines.

\subsection{Summarized Results of Adult Dataset}
\label{sec:adult}

\begin{figure}[htb]
     \centering
     \begin{subfigure}{0.22\textwidth}
         \centering
         \includegraphics[width=\textwidth]{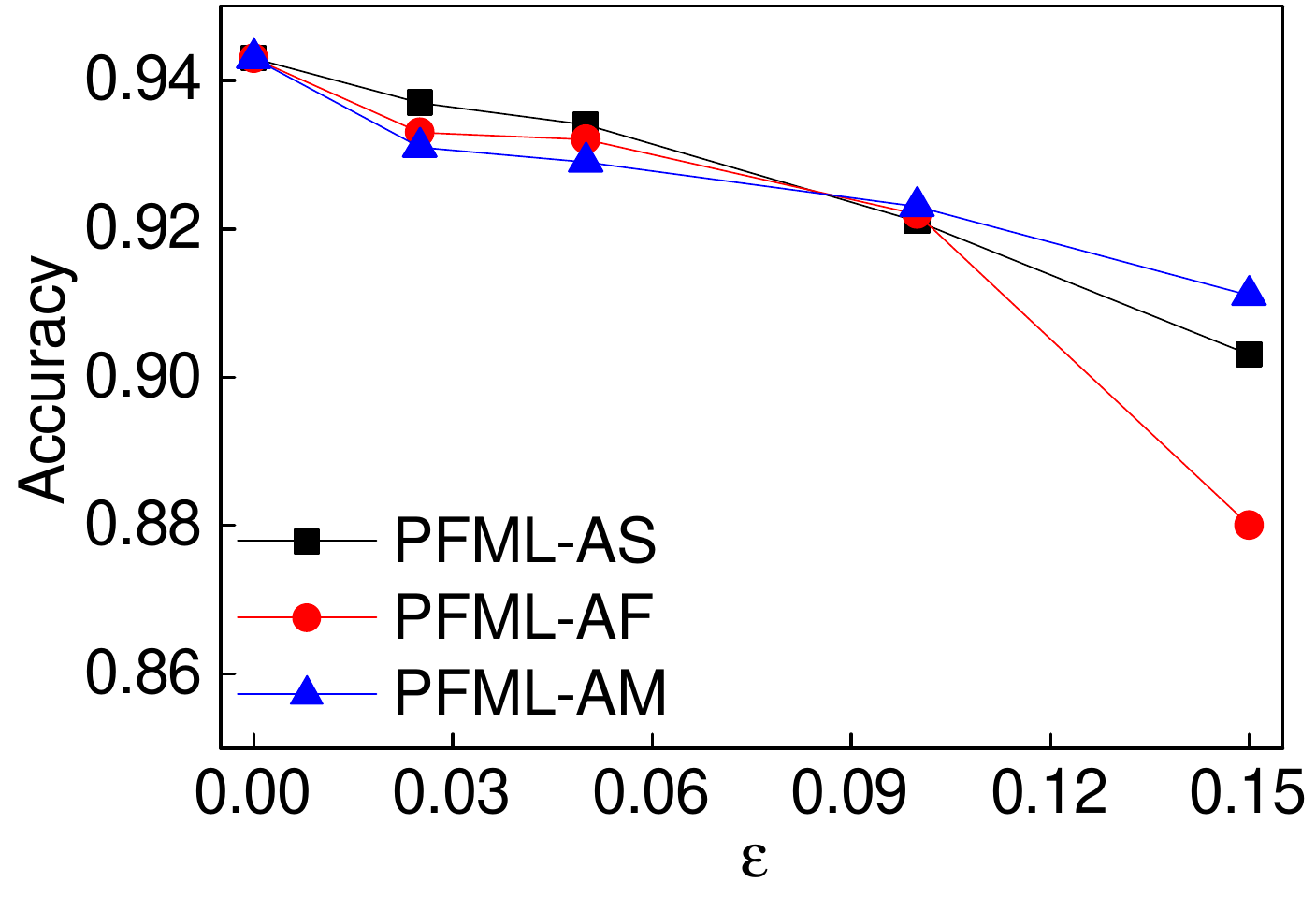}
         \caption{Accuracy}
         \label{fig:adult_acc}
     \end{subfigure}
     \hfill
     \begin{subfigure}{0.22\textwidth}
         \centering
         \includegraphics[width=\textwidth]{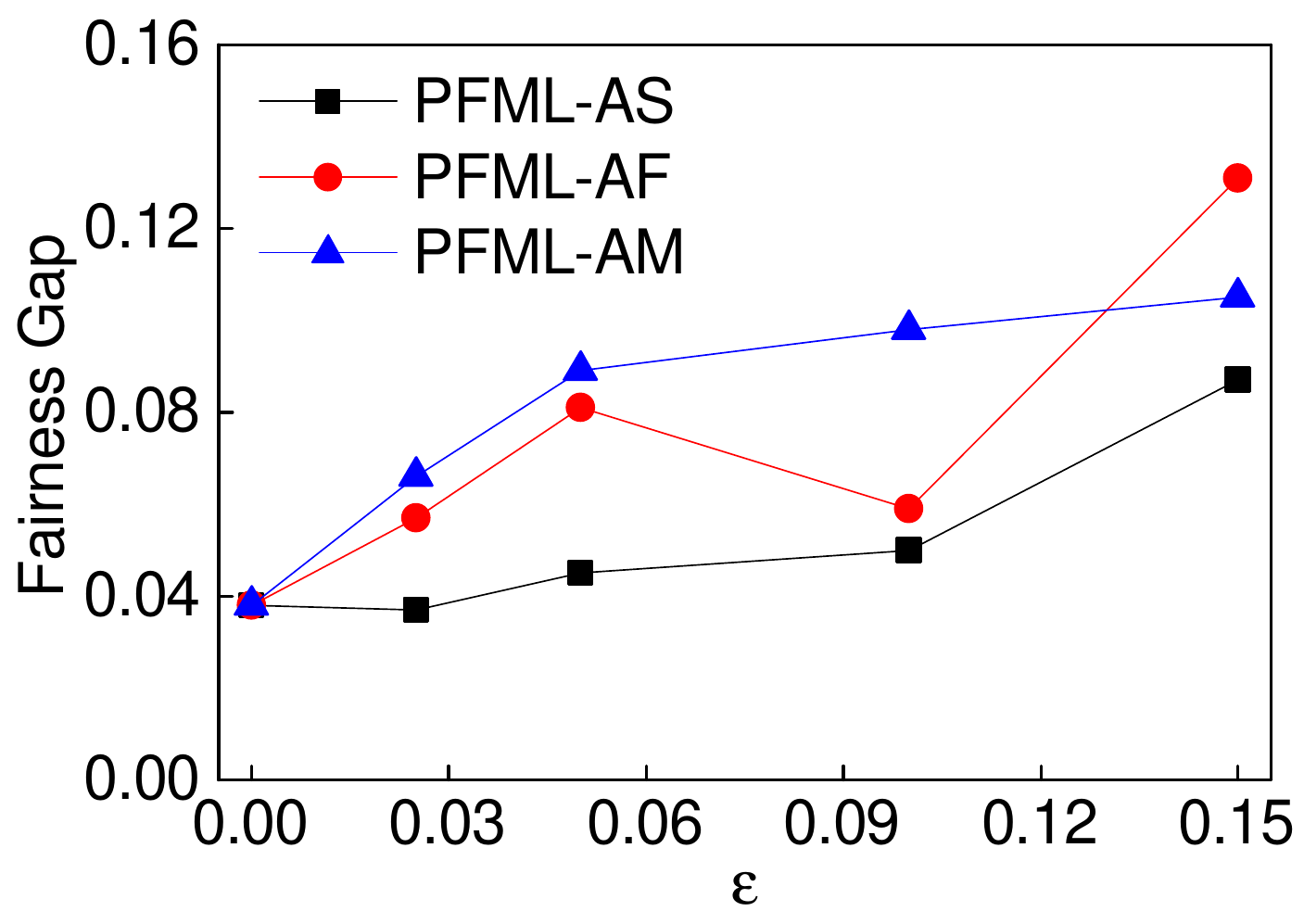}
         \caption{Fairness gap}
         \label{fig:adult_fair}
     \end{subfigure}
     \caption{Effects of ratio $\epsilon$ for fairness reduction with equalized odds (Adult).}
     \label{fig:adult}
\end{figure}

We also report our summarized results on Adult with the setting of $\delta = 0.1$, $\alpha=0.8$, and varied $\epsilon$ values. 
Figures \ref{fig:adult_acc} and \ref{fig:adult_fair} plot the curves of accuracy and fairness gap for each of three PFML attacks with the increasing $\epsilon$. The accuracy of fair reduction under PFML-AS, PFML-AF and PFML-AM attacks decreases when we increase $\epsilon$. This pattern is consistent with our observation on COMPAS. As we analyzed previously, larger $\epsilon$ indicates stronger attack as more poisoning data are injected during the model training thus cause more performance degradation. Similarly, the fairness gap under PFML-AS, PFML-AF and PFML-AM generally increases with increasing $\epsilon$.  In terms of execution time, our PFML methods take from 187.4 s to 1399.2 s with increasing $\epsilon$, which is significantly  less than the baseline models (e.g., two orders of magnitude faster than min-max attack).

\section{Related Work}
The bulk of recent research on adversarial machine learning has focused on test-time attacks where the attacker perturbs the test data to obtain a desired classification.
Designing models that are robust to such test-time attacks is an active area of research.
Train-time attacks leave the test data unchanged, and instead perturb the training data to affect the learned model. Data poisoning attacks are among the most common train-time attack methods in adversarial learning.

\citet{barreno2006can} first proposed poisoning attacks which modifies the training dataset to potentially change the decision boundaries of the targeted model.
\citet{DBLP:conf/icml/BiggioNL12} developed an approach of crafting poisoning samples  using gradient ascent.
\citet{mei2015using} developed a method that finds an optimal change to the training data when the targeted learning model is trained using a convex optimization loss and its input domain is continuous. Recent approaches include optimization-based methods \cite{koh2018stronger} (e.g., influence, KKT, and min-max), poisoning Generative Adversarial Net (pGAN) model \cite{munoz2019poisoning}, and class-oriented poisoning attacks against neural network models \cite{zhao2020class}. The influence attack is a gradient-based attack that iteratively modifies each attack sample to increase the test loss, the KKT attack selects poisoned samples to achieve pre-defined decoy parameters, and the min-max attack efficiently solves for the poisoned samples that maximize train loss as a proxy for test loss. All three attacks control both the label $y$ and input features $x$ of the poisoned points.
The pGAN model generates poisoning data points to fool the model and degrade the prediction accuracy.
Defense methods  \cite{koh2018stronger,steinhardt2017certified} typically require additional information, e.g., a labeled set of outliers or a clean set, and apply supervised classification to separate outliers from normal samples.

There have been a few works on attacking fair machine learning models very recently \cite{chang2020adversarial,solans2020poisoning,roh2020fr,DBLP:journals/corr/abs-2012-08723}.
\citet{solans2020poisoning}
developed a gradient-based poisoning attack to increase classification disparities among different groups.
\citet{DBLP:journals/corr/abs-2012-08723} also focused on attacking FML models trained with fairness constraint of demographic disparity. They developed anchoring attack and influence attack and focused on demographic disparity.
\citet{chang2020adversarial} formulated the adversarial FML as a bi-level optimization and focused on attacking FML models trained with equalized odds. To tackle the challenges of the non-convex loss functions and the non-additive function of equalized odds, they further developed two approximate algorithms.
\citet{roh2020fr} developed a GAN-based model that tries to achieve accuracy, fairness and robustness against adversary attacks.



\section{Conclusions and Future Work}
We present a poisoning sample based framework that can attack model accuracy and algorithmic fairness.  The three attacks studied in this paper are special cases of gradient-based attacks and belong to indiscriminate attacks. In our future work, we will extend our approach to other attacks, e.g.,  the targeted attacks that seek to cause errors on specific test examples, and other fairness notions, e.g., counterfactual fairness~\cite{DBLP:conf/nips/KusnerLRS17,DBLP:conf/nips/Wu0WT19}. We will also investigate robust defense approaches against attacks on fair machine learning models, e.g., by applying multi-gradient algorithms for multi-objective optimization \cite{liu2019stochastic, desideri2012multiple} and robust learning \cite{taskesen2020distributionally}.


\section{Acknowledgments}
This work was supported in part by NSF 1564250, 1564039, 1946391, and 2137335.


\end{document}